\documentclass{article}
\usepackage{spconf,amsmath,amssymb,graphicx}
\usepackage[labelfont=bf]{caption}
\usepackage{booktabs}
\usepackage{subcaption}
\usepackage{url}
\title{Decoding Functional Networks for Visual Categories via GNNs}

\name{Shira Karmi $^{\dagger,\diamond}$ \qquad Galia Avidan$^{\star,\diamond}$ \qquad Tammy Riklin Raviv$^{\dagger,\diamond}$}
\address{$^{\dagger}$ The School of Electrical and Computer Engineering, Ben-Gurion University of the Negev \\
     $^{\star}$ Psychology Department, Ben-Gurion University of the Negev\\
     $^{\diamond}$ The School of Brain Sciences and Cognition, Ben-Gurion University of the Negev}
\begin{document}
\ninept
\maketitle

\begin{abstract}
Understanding how large-scale brain networks represent visual categories is fundamental to linking perception and cortical organization. Using high-resolution 7T fMRI from the Natural Scenes Dataset, we construct parcel-level functional graphs and train a \emph{signed} Graph Neural Network that models both positive and negative interactions, with a sparse edge mask and class-specific saliency. The model accurately decodes category-specific functional connectivity states (\textit{sports}, \textit{food}, \textit{vehicles}) and reveals reproducible, biologically meaningful subnetworks along the ventral and dorsal visual pathways. This framework bridges machine learning and neuroscience by extending voxel-level category selectivity to a connectivity-based representation of visual processing.
\end{abstract}
\begin{keywords}
fMRI, Natural Scenes Dataset, Graph Neural Networks, Explainability, Functional Connectivity
\end{keywords}
\vspace{-.3cm}
\section{Introduction}
\label{sec:intro}

Representing neural activity as functional connectivity graphs derived from fMRI Blood-Oxygen-Level Dependent (BOLD) signals provides a principled way to study interactions among brain regions during perception~\cite{Bullmore2009}. Graph Neural Networks (GNNs) are well-suited for this setting because they operate directly on graph-structured data and can model higher-order patterns in functional coupling~\cite{BrainNNExplainer2024,Luo2021}.

Recent work has demonstrated the utility of GNNs for decoding cognitive states and characterizing functional organization from resting-state and clinical fMRI datasets~\cite{BrainNNExplainer2024}. However, applications of GNNs to \emph{stimulus-driven} fMRI remain comparatively limited, especially under naturalistic viewing conditions, where complex visual input elicits rich, distributed activation patterns. Interpretability is an additional challenge: while GNNs can achieve strong decoding performance, their internal decision mechanisms are often opaque. Explainable-AI (XAI) methods such as learned edge masks~\cite{Ying2019}, contrastive graph pooling~\cite{xu2024contrastive}, and Gradient--Input attribution~\cite{BrainNNExplainer2024} have been proposed, yet they are rarely combined in a unified framework for large-scale, stimulus-driven fMRI.

Here, we introduce an interpretable \emph{Signed Graph Neural Network (signed GNN)} for decoding category-specific functional connectivity during naturalistic vision. We evaluate the model on the Natural Scenes Dataset (NSD)~\cite{Allen2022NSD}, which provides high-resolution, stimulus-driven fMRI responses under richly varied visual input. Positive correlations capture cooperative co-fluctuations, whereas negative correlations capture inverse co-fluctuations between networks. By modeling both, the signed GNN represents complementary neural mechanisms underlying category-selective responses. Global interpretability is achieved through a sparsity-regularized edge mask, while class-specific relevance is obtained via Gradient–Input attribution.

Neuroscientific work has shown that visual categories evoke reliable and spatially distinct cortical responses.  Food-selective regions cluster in ventral occipitotemporal cortex and extend toward limbic–orbitofrontal areas involved  in reward and gustation~\cite{Chen2022Food,Lindquist2022Food,Jain2023}.  Sports-related perception recruits dorsal visuomotor and parietal pathways supporting motion, attention, and action processing ~\cite{Ayzenberg2022Dorsal,Wright2007Tennis}. 
Vehicle recognition engages lateral occipital and posterior ventral visual areas associated with object shape and  viewpoint representations~\cite{andresen2008viewpoint}. 
Beyond local selectivity, a large body of functional connectivity research has demonstrated that visual processing  relies on distributed, multi-region coupling patterns across ventral, dorsal, and higher-order association networks 
(e.g.,~\cite{Hutchison2013Dynamic,Baldassano2017Story,Haxby2020Hyper}). 
However, most connectivity studies rely on correlation matrices or statistical comparisons rather than learned, explainable models.
We focus on \emph{sports}, \emph{food}, and \emph{vehicle} images because they provide 
reliable class distinctions in natural scenes and engage broad, multi-region cortical systems. These categories span complementary large-scale networks, making them well suited for studying how category information is encoded in distributed functional connectivity.

Overall, this work advances explainable graph learning for stimulus-driven fMRI by integrating large-scale naturalistic data with an interpretable signed GNN framework. Our contributions are threefold:
(1) a multimodal labeling pipeline that fuses COCO instance masks and captions to derive reliable super-category assignments;
(2) a signed GNN that models both positive and negative connectivity and provides global and class-specific explanations via a learned edge mask and Gradient–Input relevance; and
(3) a connectivity-based analysis that uncovers category-selective subnetworks spanning ventral, dorsal, and limbic pathways.
Taken together, these components form a unified approach for characterizing how naturalistic visual input shapes distributed functional interactions in human cortex.
\vspace{-.3cm}
\section{Methods}
\vspace{-.3cm}
\label{sec:methods}
Our goal is decoding category-specific functional connectivity states using an interpretable GNN framework that links aggregated stimulus-evoked brain responses to semantic categories. The approach is modular and applicable to stimulus-based fMRI datasets with regional time series.

We denote the parcellated fMRI time series of subject $s$ by
$\mathbf{X}_s \in \mathbb{R}^{N \times P \times T}$, where $N$ is the number of stimulus presentations, $P$ is the number of
cortical parcels and $T$ the number of timepoints per trial. 
Our goal is to classify category-specific connectivity states into one of $C$ super-categories using an interpretable signed GNN.

For network-level analysis, each block is represented by a functional
connectivity matrix $A \in \mathbb{R}^{P \times P}$ capturing pairwise
parcel correlations. We decompose each matrix into positive and negative
components $(A^{+}, A^{-})$, which together form the signed input graph
processed by the GNN.

The methodological objectives are therefore threefold:
(1)~to derive reliable semantic category labels for each stimulus using a
multimodal annotation-fusion scheme;
(2)~to learn category-discriminative representations of signed functional
connectivity using a signed GNN; and
(3)~to obtain interpretable explanations of the connectivity structure
underlying each visual category.

%The rest of this section describes the dataset
%(Section~\ref{sec:dataset}), preprocessing and category %construction
%(Section~\ref{sec:preprocessing}), connectivity %formulation
%(Section~\ref{sec:conn_mat}), the signed GNN %architecture
%(Section~\ref{sec:signedgnn}), and the interpretability %methods
%(Section~\ref{sec:explainability}).
\vspace{-.3cm}
\subsection{Dataset}
\label{sec:dataset}
We used the Natural Scenes Dataset (NSD) \cite{Allen2022NSD}, a large-scale
7~Tesla fMRI resource providing high-resolution cortical responses to
naturalistic stimuli. Across eight participants, the experiment presented up to
10{,}000 unique images drawn from the \emph{Microsoft COCO} dataset
\cite{Lin2015COCO}, which contains complex, multi-object scenes.

Each image was displayed for 4\,s while participants performed a continuous
recognition task, indicating whether it had appeared earlier in the experiment.
This paradigm maintains attention and minimizes habituation across thousands of
trials.
NSD data were acquired at 1.8\,mm isotropic resolution and projected onto each
participant’s cortical surface using subject-specific reconstructions, yielding
high-fidelity, surface-based fMRI aligned to cortical parcels. This makes NSD
well suited for connectivity analysis and graph-based modeling.
\vspace{-.3cm}
\subsection{Preprocessing}
\label{sec:preprocessing}
% \begin{figure}[t]
%   \centering
%   \includegraphics[width=1\linewidth]{images/gnn.png}
%   \caption{An overview of our proposed framework}
%   \label{fig:gnn}
% \end{figure}
All fMRI data were preprocessed using the NSD-derived routines from the
Algonauts 2023 pipeline \cite{Algonauts2023Prep}, including motion correction,
detrending, hemodynamic normalization, surface projection, and standardized
extraction of parcel-wise responses. Each subject’s cortical surface was then
parcellated into $P$ regions using the multimodal HCP--MMP1 atlas
\cite{Glasser2016}, which provides neurobiologically consistent boundaries.

\begin{figure}[t]
    \centering

    \begin{minipage}[b]{0.31\linewidth}
        \centering
        \includegraphics[width=\linewidth]{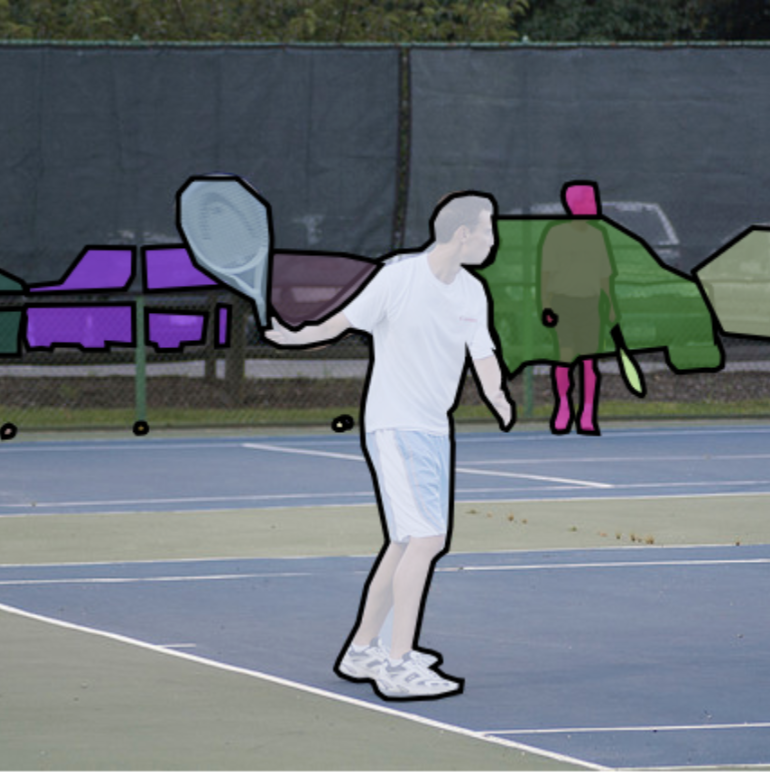}
        \caption*{(a) Sport}
    \end{minipage}
    \hfill
    \begin{minipage}[b]{0.32\linewidth}
        \centering
        \includegraphics[width=\linewidth]{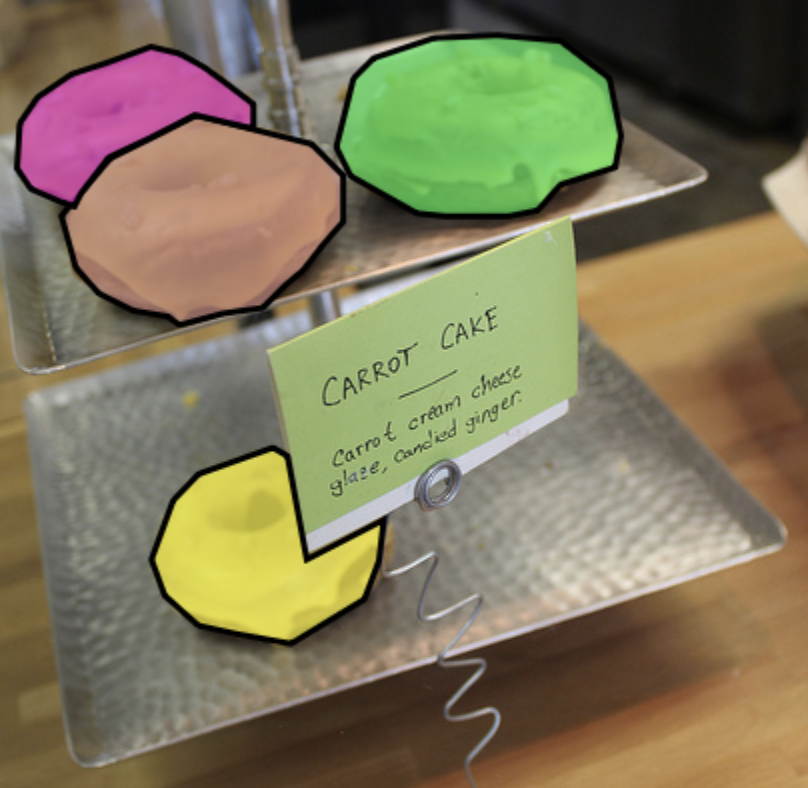}
        \caption*{(b) Food}
    \end{minipage}
    \hfill
    \begin{minipage}[b]{0.32\linewidth}
        \centering
        \includegraphics[width=\linewidth]{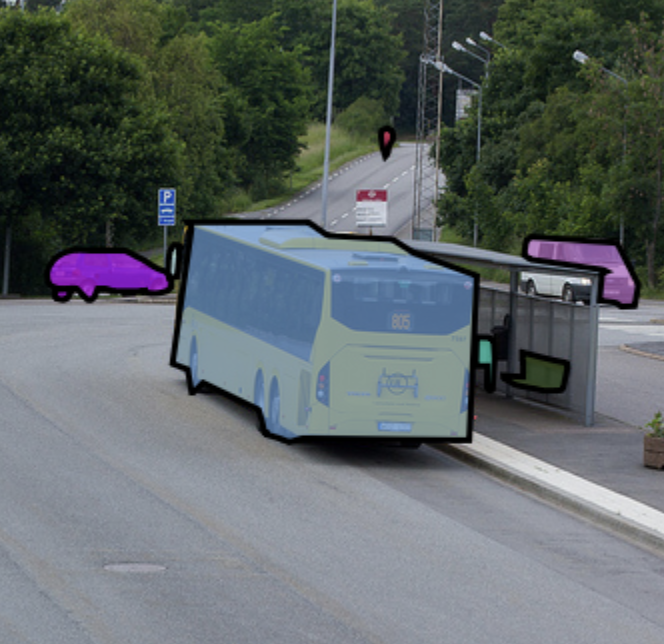}
        \caption*{(c) Vehicle}
    \end{minipage}

    \caption{Example NSD stimulus with COCO instance masks and caption-derived category evidence for\textit{sports}, \textit{food} and \textit{vehicle}}
    \label{fig:mask_caption_example}
\end{figure}

For each subject, we extracted parcel-averaged time series across all runs,
yielding denoised and standardized regional signals suitable for
connectivity-based analysis.
\\
\textbf{Category Fusion.}
We define three super-categories—\emph{sports}, \emph{food}, and
\emph{vehicle}—and derive multimodal evidence for each NSD image using its
COCO~2017 instance masks and captions. For each category $c$ and image $i$,
mask evidence $\mathrm{mask}_{c}(i)$ quantifies the fraction of the image
occupied by objects belonging to $c$, and text evidence $\mathrm{text}_{c}(i)$
captures the relative frequency of category-related terms in the caption. These
signals are combined using a category-specific weight $\alpha_c$ to yield
$
\mathrm{score}_{c}(i)
= \alpha_{c}\,\mathrm{text}_{c}(i)
 + (1{-}\alpha_{c})\,\mathrm{mask}_{c}(i).
$
An illustration of this multimodal fusion appears in
Fig.~\ref{fig:mask_caption_example}.

% \begin{figure}[t]
%   \centering
%   \includegraphics[width=0.3\linewidth]{images/mask_caption_example.png}
%   \caption{Example NSD stimulus with COCO instance masks and caption-derived category evidence for \textit{sports} and \textit{food}.}
%   \label{fig:mask_caption_example}
% \end{figure}

\textbf{Category Assignment and Duplication.}
Since natural scenes often contain multiple semantic elements, an image may
provide meaningful evidence for more than one super-category. To convert the
continuous fusion scores into training samples, we discretize each score into
approximately 10\% units. For each category $c$, the number of samples
generated by image $i$ is
$
n_{c}(i)=\left\lfloor\,10\cdot \mathrm{score}_{c}(i)\,\right\rceil ,
$
so images with stronger category evidence contribute proportionally more
examples while keeping class sizes balanced. A global train/validation/test
split is defined on the image indices \emph{before} duplication to avoid
leakage across samples or categories.
\vspace{-.3cm}
\subsection{Connectivity Matrix Construction}
\label{sec:conn_mat}

For each subject $s$, let $X_{s}^{(k)} \in \mathbb{R}^{P \times T}$ denote the
parcel-wise fMRI time series for trial $k$. Trials are grouped into blocks of
fixed size $K$. For a block containing trials
$\{X_{s}^{(k_{1})},\ldots,X_{s}^{(k_{K})}\}$, we concatenate their time series to improve SNR for connectivity estimation,
\[
\tilde{X}_{s}
=
\big[\,X_{s}^{(k_{1})}~|~X_{s}^{(k_{2})}~|~\cdots~|~X_{s}^{(k_{K})}\,\big]
\in \mathbb{R}^{P \times (KT)},
\]
and z-score the resulting matrix across time.

We compute the Pearson-correlation matrix $A$ from $\tilde{X}_{s}$, clipping extreme
values to the Fisher-consistent range $[-1,1]$. Each block yields one
connectivity matrix, producing a single graph sample. Matrices are constructed
independently for each super-category and balanced across classes for training.

To form signed graphs, we separate positive and negative correlations into two
channels,
\[
A^{+}_{ij}=\max(A_{ij},0),
\qquad
A^{-}_{ij}=\max(-A_{ij},0),
\]
which together serve as the input to the signed GNN.
Each graph represents a category-conditioned, block-level connectivity estimate, not a single-trial response.

\subsection{Signed GNN Architecture}
\label{sec:signedgnn}

Each connectivity graph is encoded as a two-channel adjacency tensor
$A = (A^{+},A^{-}) \in \mathbb{R}^{2 \times P \times P}$, where $A^{+}$ and
$A^{-}$ store the positive and negative correlations, respectively.

The interpretation of negative correlations in fMRI remains debated, as these
values are typically weaker and can be affected by preprocessing choices such as
global signal regression~\cite{Murphy2009GSR}. We therefore treat $A^{-}$ not as
a biologically inhibitory signal but simply as an additional relational feature
capturing inverse co-fluctuations that may aid classification.

Separating positive and negative edges allows the signed GNN to learn
data-driven distinctions between the two without imposing assumptions about
their physiological meaning. This representation preserves the full signed
connectivity structure and enables the model to emphasize or suppress negative
edges depending on their utility for category decoding.
\\
\textbf{Learnable edge mask.}
To promote sparse and interpretable connectivity patterns, we introduce a shared symmetric mask
$M \in [0,1]^{P \times P}$ that multiplicatively gates both adjacency channels,
\[
\widetilde{A}^{+} = A^{+} \odot M, \qquad
\widetilde{A}^{-} = A^{-} \odot M.
\]
The mask allows the network to retain, attenuate, or suppress specific
connections. To encourage a small set of informative edges, we regularize $M$
with sparsity and near-binary constraints:
\[
\mathcal{L}_{\text{mask}}
= \lambda_{1}\|M\|_{1}
+ \lambda_{e}\, M \odot (1{-}M),
\]
promoting sparse, interpretable connectivity patterns shared across categories.
\\
\textbf{Signed graph convolutions.}
The model contains two signed graph-convolution layers. Let
$h_i^{(\ell)}$ denote the feature vector of node $i$ at layer $\ell$. Each layer
updates node features via
\[
h_i^{(\ell+1)} = \sigma\!\left(
W_{+}^{(\ell)}\!\sum_{j} \widetilde{A}^{+}_{ij} h_j^{(\ell)}
+
W_{-}^{(\ell)}\!\sum_{j} \widetilde{A}^{-}_{ij} h_j^{(\ell)}
+ b^{(\ell)}
\right),
\]
where \(W_{+}^{(\ell)}\) and \(W_{-}^{(\ell)}\) learn from positive and negative
interactions, respectively. Node features are initialized as one-hot identity
vectors, ensuring each parcel begins with a distinct representation.
\\
\textbf{Pooling and classification.}
Node embeddings after the second convolutional layer are aggregated by
global mean pooling:
$
g = \frac{1}{P}\sum_{i=1}^{P} h_i^{(2)}.
$
The pooled representation is passed through a  multilayer perceptron:
\[
z = \phi(W_{1} g + b_{1}), \qquad
\hat{y} = \mathrm{softmax}(W_{2} z + b_{2}),
\]
producing class probabilities for each of the categories.
\\
\textbf{Interpretability outputs.}
The mask $M$ gates all adjacency operations, thus, it yields a
global, category-agnostic measure of connectivity importance.
Class-specific explanations are obtained via gradient--input attribution
(Section~\ref{sec:explainability}),
providing fine-grained maps of how individual connections influence each
category decision.
\vspace{-.3cm}
\subsection{Explainability}
\label{sec:explainability}
To interpret how the signed GNN uses functional connectivity to perform 
category decoding, we combine two complementary forms of analysis:
(i)~a global, sample-agnostic edge mask that highlights sparse connectivity patterns shared across the dataset~\cite{Ying2019}, and 
(ii)~class-specific gradient-input attribution that identifies which edges
drive each category decision~\cite{shrikumar2017learning,simonyan2014deep}.
\\
\textbf{(i) Global edge-importance mask.}
Because the learnable mask $M$ multiplicatively gates all adjacency matrices
(Section~\ref{sec:methods}), the entry $M_{ij}$ directly reflects the 
importance of the connection between parcels $i$ and $j$ across 
\emph{all} categories and samples.
After training, $M$ highlights sparse, interpretable connectivity patterns shared across categories. 
We visualize:
$
M_{\text{sym}} = \frac{1}{2}(M + M^{\top}),
$
and interpret high-valued edges as the core connections consistently used
during graph convolution.  
Parcel-level global relevance is computed as the weighted node degree:
\[
r^{\text{global}}_i = \sum_{j} M_{\text{sym},ij}.
\]
\textbf{(ii) Class-specific saliency maps.}
To capture category-dependent connectivity patterns, we compute
gradient--input attribution for each class $c\in\{\text{sports},\text{food},\text{vehicle}\}$.
Given the class logit $f_c(A)$ and original adjacency channel 
$A^\pm$, the class-specific relevance is:
\[
S_{c}^{\pm} = 
\left| 
\frac{\partial f_c}{\partial A^{\pm}} \odot A^{\pm}
\right|,
\]
followed by channel fusion:
$S_{c} = S_{c}^{+} + S_{c}^{-}.$
We symmetrize and zero the diagonal,
$\widetilde{S}_{c} = \mathrm{sym}(S_{c}) - \mathrm{diag}(S_{c}),$
and select the top-$k$ edges by magnitude.
Node-level class-specific relevance is then computed as:
$r^{(c)}_i = \sum_{j} \widetilde{S}_{c,ij}.$\\
\textbf{Spatial visualization.}
The relevance vectors $r^{\text{global}}$ and $r^{(c)}$ 
are mapped onto the cortical surface using the Human Connectome Project multi-modal parcellation 1.0 (HCP-MMP1) \cite{Glasser2016} parcellation,
yielding interpretable region-level importance maps.  
The global mask highlights a shared connectivity scaffold, whereas
class-specific saliency maps reveal distinct subnetworks
supporting \textit{sports}, \textit{food}, and \textit{vehicle} recognition.
\vspace{-.3cm}
\section{Experiments}
\label{sec:experiments}
\vspace{-.3cm}
\subsection{Hemisphere Selection}
We apply the full pipeline separately to the right (RH) and left (LH)
hemispheres. All development is performed on the RH and then transferred
unchanged to the LH, enabling controlled comparison of category-selective
connectivity patterns and assessment of hemispheric asymmetries. RH results are reported in (Table~\ref{tab:ap_by_subject_updated}) and LH
results compare to RH report in (Section~\ref{sec:LRH}).
\vspace{-.3cm}
\subsection{Setup and Data Splits}
All analyses are performed independently for each of the eight NSD
participants. To prevent data leakage, the global train/validation/test split
is defined at the \emph{image} level before any fMRI processing. Each image
contributes a variable number of samples to its assigned super-category
(\textit{sports}, \textit{food}, \textit{vehicle}) based on its multimodal
fusion score (Section~\ref{sec:methods}), with roughly one sample added for
every 10\% of category evidence.
\vspace{-.3cm}
\subsection{Training Protocol}
Models are trained independently for each subject using minibatches of
connectivity matrices derived from the training images. The signed GNN is
optimized with class-weighted cross-entropy (label smoothing 0.1) and AdamW,
with early stopping based on validation loss. Training uses a batch size of 32,
an initial learning rate of $3\times10^{-4}$, and weight decay of $10^{-3}$.
Random seeds are fixed across all preprocessing and training steps to ensure
full reproducibility.
\vspace{-.3cm}
\subsection{Evaluation}
Performance is evaluated on the held-out test split using classification
accuracy and per-class Average Precision (AP). Explainability is computed only
after model selection, with saliency and mask-derived relevance maps evaluated
per category and aggregated across subjects for group-level analysis.
\vspace{-.3cm}
\section{Results}
\label{sec:results}

\begin{figure}[htb]
\begin{minipage}[b]{0.48\linewidth}
  \includegraphics[width=4.0cm]{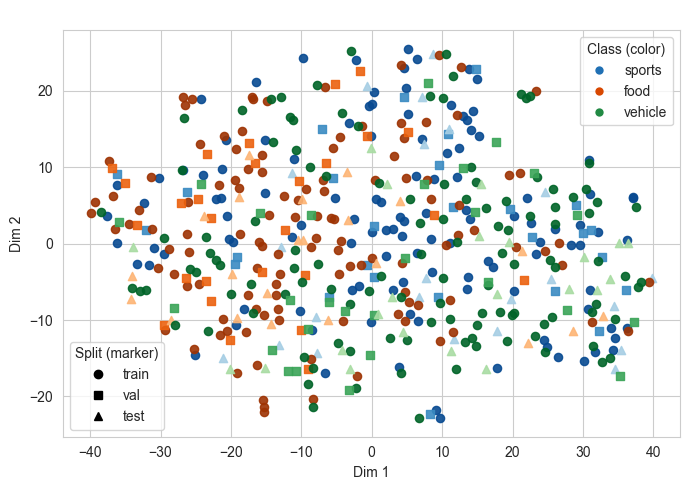}
  \centerline{(a) Epoch 0}
\end{minipage}
\hfill
\begin{minipage}[b]{0.48\linewidth}
  \includegraphics[width=4.0cm]{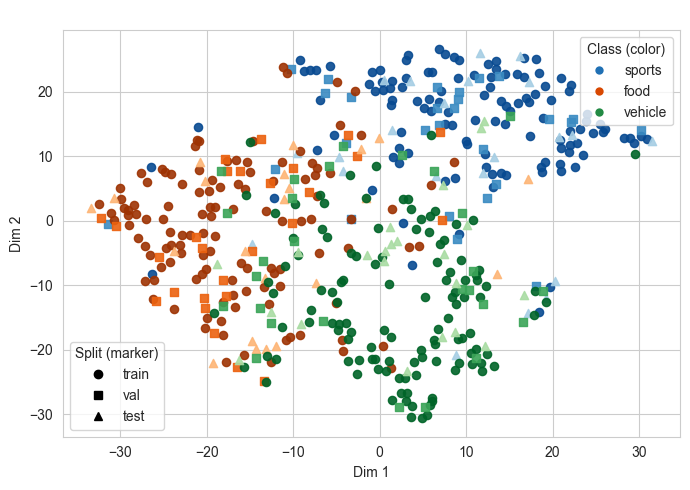}
  \centerline{(b) 1/3 training}
\end{minipage}

\begin{minipage}[b]{0.48\linewidth}
  \includegraphics[width=4.0cm]{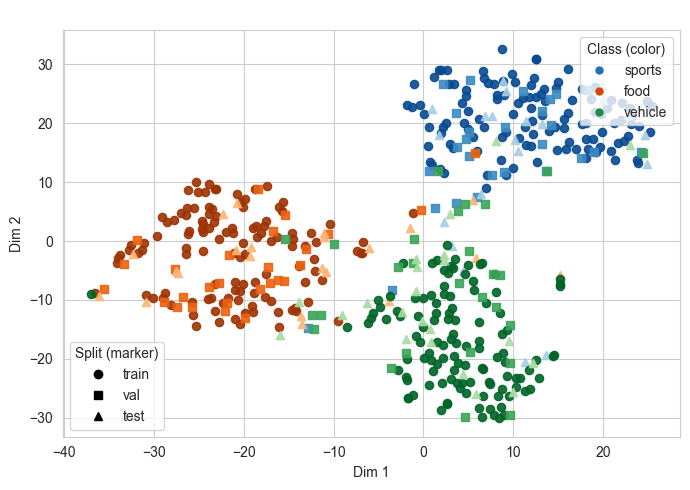}
  \centerline{(c) 2/3 training}
\end{minipage}
\hfill
\begin{minipage}[b]{0.48\linewidth}
  \includegraphics[width=4.0cm]{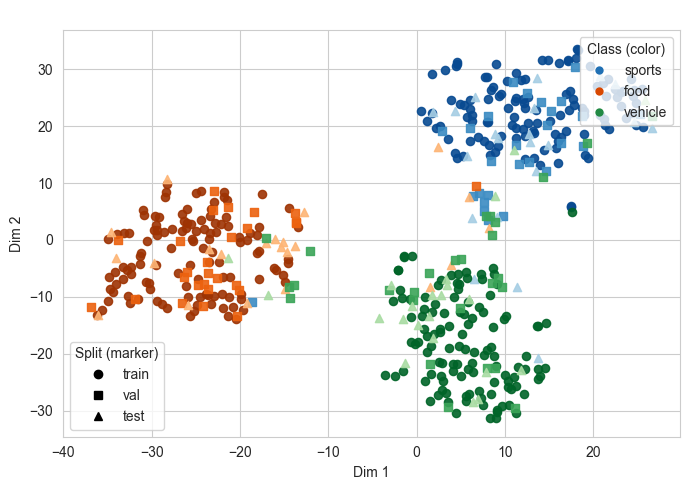}
  \centerline{(d) End of training}
\end{minipage}

\caption{\small
t-SNE progression of graph embeddings across training.
Clusters emerge for \textit{sports} (blue), \textit{food} (red), and
\textit{vehicle} (green), showing the signed GNN gradually disentangles
category-selective connectivity patterns.}
\label{fig:tsne_progression}
\end{figure}
\vspace{-.3cm}
\subsection{Latent-Space Organization}
Projecting pooled embeddings to 2D with t-SNE
(Fig.~\ref{fig:tsne_progression}) shows that category structure is absent at
initialization but becomes increasingly separable during training. By
convergence, \textit{sports}, \textit{food}, and \textit{vehicle} occupy
distinct regions of latent space, indicating that the model learns
category-discriminative connectivity features. Trajectories are consistent
across participants, suggesting a shared representational geometry.
\vspace{-.3cm}
\subsection{Quantitative Evaluation}
Training converges within 50--60 epochs with minimal overfitting. Subjects show
right-hemisphere (RH) accuracy \textbf{0.78} and macro-AP \textbf{0.88}
(Table~\ref{tab:ap_by_subject_updated}). \textit{Food} yields the highest
category AP (\textbf{0.92}), followed by \textit{sports} (\textbf{0.88}) and
\textit{vehicle} (\textbf{0.84}). Performance is comparable on the left
hemisphere (LH; accuracy \textbf{0.80}, macro-AP \textbf{0.91}), with the same
category ranking. Inter-subject variability is modest and within expected
ranges for fMRI data.
\begin{table}[t]
\centering
\caption{Per-subject test accuracy and Average Precision (AP). Macro-AP is the
mean AP across the three classes. Final rows show mean $\pm$ standard
deviation.}
\label{tab:ap_by_subject_updated}
\begin{tabular}{lccccc}
\hline
\textbf{Subj.} & \textbf{Acc.} & \textbf{Sports} & \textbf{Food} & \textbf{Veh.} & \textbf{Macro-AP} \\
\hline
1 & 0.790 & 0.916 & 0.939 & 0.829 & 0.895 \\
2 & 0.769 & 0.860 & 0.930 & 0.837 & 0.876 \\
3 & 0.640 & 0.739 & 0.843 & 0.643 & 0.742 \\
4 & 0.870 & 0.923 & 0.942 & 0.880 & 0.915 \\
5 & 0.810 & 0.929 & 0.949 & 0.918 & 0.932 \\
6 & 0.818 & 0.950 & 0.959 & 0.910 & 0.940 \\
7 & 0.790 & 0.854 & 0.961 & 0.804 & 0.873 \\
8 & 0.746 & 0.870 & 0.859 & 0.858 & 0.863 \\
\hline
\textbf{Mean} & \textbf{0.78} & \textbf{0.88} & \textbf{0.92} & \textbf{0.84} & \textbf{0.88} \\
$\pm$ STD & $\pm$0.06 & $\pm$0.07 & $\pm$0.04 & $\pm$0.09 & $\pm$0.06 \\
\hline
\end{tabular}
\end{table}
\vspace{-.3cm}
\subsection{Category-Selective Connectivity Networks}
\begin{figure}[t]
  \centering
  \includegraphics[width=\linewidth]{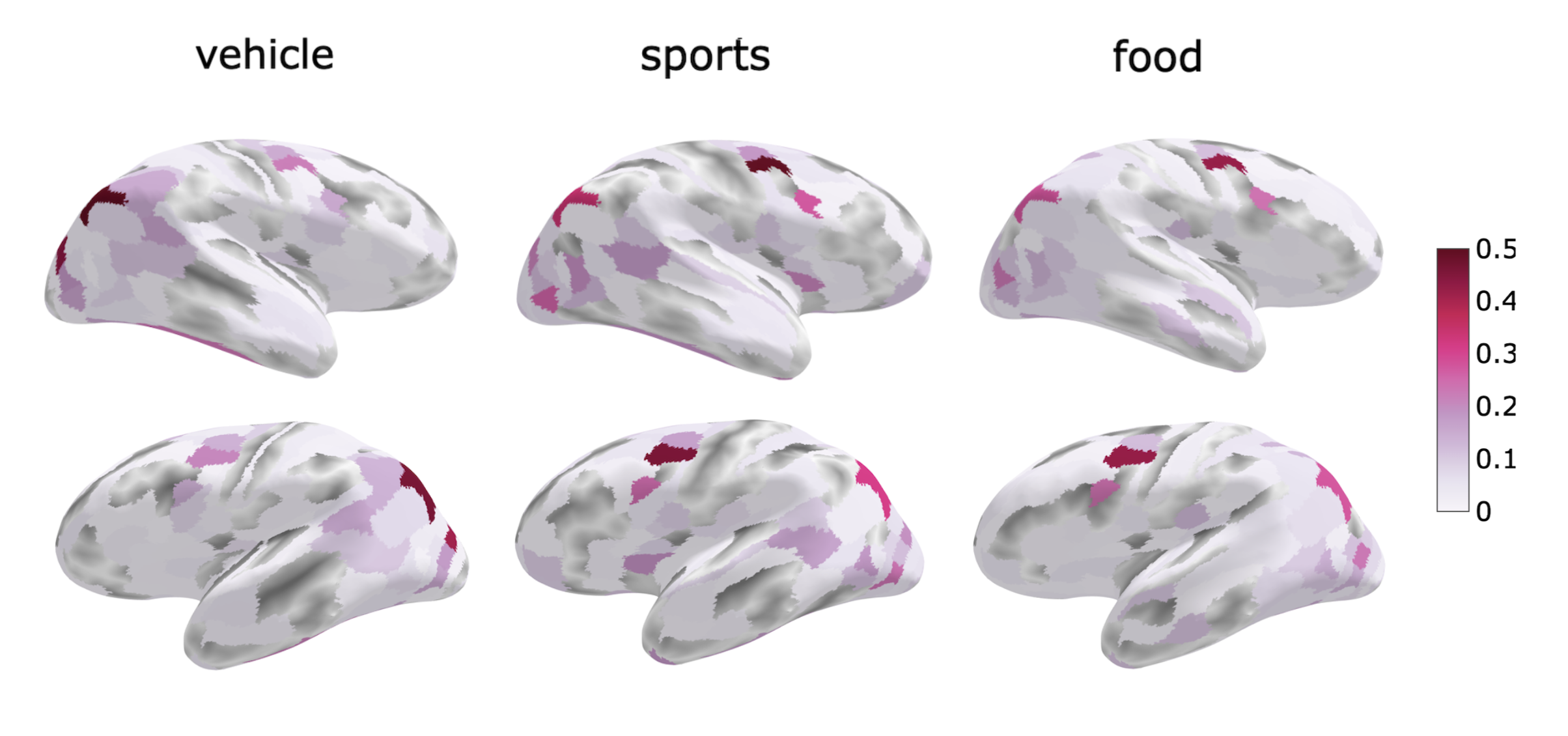}
  \caption{Top-weighted parcels for sports, food, and vehicle.
  Warmer colors denote higher relevance from the learned mask and saliency.}
  \label{fig:brain}
\end{figure}
Relevance maps derived from the mask and saliency reveal coherent,
category-specific subnetworks (Fig.~\ref{fig:brain}). Ventral visual parcels
(VMV1–3, FFC) contribute across all categories, while dorsal visuomotor and
limbic regions differentiate them:
\\
\noindent\textbf{Food.}
High relevance appears in ventral occipitotemporal areas
(\textit{VMV1–2}, \textit{FFC}) and limbic–insular parcels
(\textit{FOP4}, \textit{PHA2}), consistent with reports of food-selective
ventral cortex~\cite{Chen2022Food,Lindquist2022Food,Jain2023}.
\\
\noindent\textbf{Sports.}
Dorsal visuomotor parcels (\textit{MT/MST}, \textit{V7},
\textit{VIP/LIP}, \textit{7Am}) dominate, aligning with dorsal-stream roles
in motion and action processing~\cite{Ayzenberg2022Dorsal,Wright2007Tennis}.
\\
\noindent\textbf{Vehicles.}
Top-weighted parcels cluster in ventral and lateral occipitotemporal cortex
(\textit{LO1–3}, \textit{VMV1–3}) with contributions from posterior parietal
areas (\textit{PGs}), consistent with object form and layout processing and
viewpoint-invariant recognition~\cite{andresen2008viewpoint}.

Interpretation relies on the HCP--MMP1 atlas~\cite{Glasser2016}, confirming
that the signed GNN recovers subnetworks aligned with established ventral,
dorsal, and limbic processing pathways.
\vspace{-.3cm}
\subsection{Cross-Subject Consistency}
Category-selective subnetworks are reproducible across participants, with high
pairwise correlations of node-importance maps. Mild differences (e.g.,
weaker ventral–limbic expression for food in some subjects) show no systematic
pattern and likely reflect typical fMRI variability.
\vspace{-.3cm}
\subsection{Comparison to Contrastive Learning}
A GraphCL-based contrastive baseline \cite{You2020GraphCL} performs substantially worse than the
supervised signed GNN (accuracy $0.637$ vs.\ $0.780$; macro-AP $0.698$ vs.\
$0.880$; Table~\ref{tab:mean_comparison_transposed}). Statistical tests confirm a
consistent subject-wise advantage (Wilcoxon $p=0.0078$; $t$-test
$p=2.8\times10^{-5}$), reflecting the importance of supervision for capturing
category-specific signed connectivity patterns.

\begin{table}[t]
\centering
\caption{Mean performance of contrastive learning vs.\ signed GNN.}
\label{tab:mean_comparison_transposed}
\begin{tabular}{lccccc}
\hline
\textbf{Method} & \textbf{Acc.} & \textbf{Sports} & \textbf{Food} & \textbf{Veh.} & \textbf{Macro-AP} \\
\hline
CL   & 0.637 & 0.728 & 0.752 & 0.613 & 0.698 \\
Our  & 0.780 & 0.880 & 0.920 & 0.840 & 0.880 \\
\hline
\end{tabular}
\end{table}
\vspace{-.3cm}
\subsection{Right–Left Hemisphere Comparison}
\label{sec:LRH}
LH performance closely matches RH, and connectivity explanations show
broad agreement for \textit{sports} and \textit{vehicle}. Food-selective maps
exhibit stronger, more coherent ventral–limbic involvement in RH, consistent
with prior NSD reports \cite{Jain2023}.
\vspace{-.3cm}
\subsection{Methodological Note and Limitations}
The signed GNN provides an interpretable framework for modeling signed functional connectivity through a combination of a learnable global mask and class-specific saliency. The analysis focuses on stable, category-selective connectivity configurations rather than single-trial or time-resolved dynamics. Limitations include label ambiguity in COCO-derived categories, hemisphere-only analyses, and the use of static correlations; aggregation across samples mitigates label noise.

\vspace{-.3cm}
\section{Conclusion}
This study demonstrates that Signed Graph Neural Networks offer a powerful and interpretable way to decode large-scale fMRI connectivity during natural image viewing. By combining a learned edge mask with gradient-based saliency, the model reveals clear, reproducible subnetworks for \textit{food}, \textit{sports}, and \textit{vehicle}—closely aligning with well-established ventral and dorsal visual hierarchies. Together, these findings provide compelling, population-level evidence for how the human brain organizes and distributes semantic information under naturalistic conditions.

Future work will jointly model both hemispheres to examine cross-hemispheric interactions and lateralization, and will explore individual differences in semantic preferences and behavior. More broadly, interpretable graph models offer a scalable framework for mapping how distributed neural systems represent semantic and affective information during real-world perception.
\section{Compliance with Ethical Standards}
This study was conducted using publicly available human fMRI data from the Natural Scenes Dataset (NSD)~\cite{Allen2022NSD}. 
Ethical approval for data collection was obtained by the original investigators at the University of Minnesota under Institutional Review Board (IRB) protocol number 00005568, and all participants provided informed consent. 
The present analysis used only anonymized, open-access data and therefore required no additional ethical approval.

\section{Acknowledgments}
This work was supported by the School of Electrical and Computer Engineering and the Zelman Center for Brain Science Research at Ben-Gurion University of the Negev through departmental and research scholarships awarded. 
The authors declare no conflicts of interest.
\bibliographystyle{IEEEbib}
\bibliography{refs}

@Article{Allen2022NSD,
  author    = {E.~J. Allen and G. St-Yves and Y. Wu and J.~L. Breedlove and J.~S. Prince and L.~T. Dowdle and M. Nau and B. Caron and F. Pestilli and I. Charest and J.~B. Hutchinson and T. Naselaris and K. Kay},
  title     = {A massive 7 {T} {fMRI} dataset to bridge cognitive neuroscience and artificial intelligence},
  journal   = {Nature Neuroscience},
  year      = {2022},
  volume    = {25},
  pages     = {116--126},
  doi       = {10.1038/s41593-021-00962-x}
}

@Article{BrainNNExplainer2024,
  author    = {X. Zhou and Y. Li and L. He and others},
  title     = {Brain{NNE}xplainer: Interpretable Graph Neural Networks for Brain Connectivity Analysis},
  journal   = {Medical Image Analysis},
  year      = {2024}
}

@Article{Chen2022Food,
  author    = {Y. Chen and M. Khosla and N.~A. R. Murty and N. Kanwisher},
  title     = {A highly selective response to food in human visual cortex revealed by hypothesis-free voxel decomposition},
  journal   = {Current Biology},
  year      = {2022},
  volume    = {32},
  number    = {13},
  pages     = {2875--2883.e6},
  doi       = {10.1016/j.cub.2022.05.045}
}

@Article{Jain2023,
  author    = {N. Jain and A. Wang and M.~M. Henderson and others},
  title     = {Selectivity for food in human ventral visual cortex},
  journal   = {Communications Biology},
  year      = {2023},
  volume    = {6},
  pages     = {175}
}

@Article{Lindquist2022Food,
  author    = {I.~M. L. Pennock and C. Cziraki and M. Brett and N. Kriegeskorte and N. Rezaii and M. C. Iordan and others},
  title     = {Color-biased regions in the ventral visual pathway are food selective},
  journal   = {Current Biology},
  year      = {2023},
  volume    = {33},
  number    = {1},
  pages     = {134--146},
  doi       = {10.1016/j.cub.2022.10.005}
}

@Article{Xu2024Contrastive,
  author    = {J. Xu and Q. Bian and X. Li and A. Zhang and Y. Ke and W. K. J. Sim and B. Gulyás},
  title     = {Contrastive Graph Pooling for Explainable Classification of Brain Networks},
  journal   = {IEEE Transactions on Medical Imaging},
  year      = {2024}
}

@InProceedings{Ying2019,
  author    = {R. Ying and D. Bourgeois and J. You and M. Zitnik and J. Leskovec},
  title     = {{GNN}Explainer: Generating Explanations for Graph Neural Networks},
  booktitle = {Advances in Neural Information Processing Systems (NeurIPS)},
  organization = {IEEE},
  year      = {2019},
  pages     = {9240--9251}
}

@Article{Ayzenberg2022Dorsal,
  author    = {V. Ayzenberg and T. Stephens and S. F. Lourenco},
  title     = {The dorsal visual pathway represents object-centered part relations},
  journal   = {Journal of Neuroscience},
  year      = {2022},
  volume    = {42},
  number    = {23},
  pages     = {4693--4705},
  doi       = {10.1523/JNEUROSCI.1963-21.2022}
}

@Article{Wright2007Tennis,
  author    = {M.~J. Wright and R.~C. Jackson},
  title     = {Brain regions concerned with perceptual skills in tennis: an {fMRI} study},
  journal   = {International Journal of Psychophysiology},
  year      = {2007},
  volume    = {63},
  number    = {2},
  pages     = {214--220},
  doi       = {10.1016/j.ijpsycho.2006.03.018}
}

@Article{Andresen2008Viewpoint,
  author    = {D.~R. Andresen and others},
  title     = {The representation of object viewpoint in human visual cortex},
  journal   = {NeuroImage},
  year      = {2008},
  volume    = {40},
  number    = {1},
  pages     = {198--210}
}

@Article{Glasser2016,
  author    = {M.~F. Glasser and T.~S. Coalson and E. C. Robinson and C. D. Hacker and J. Harwell and E. Yacoub and K. Ugurbil and J. Andersson and C. F. Beckmann and M. Jenkinson and S. M. Smith and D. C. Van Essen},
  title     = {A multi-modal parcellation of human cerebral cortex},
  journal   = {Nature},
  year      = {2016},
  volume    = {536},
  number    = {7615},
  pages     = {171--178}
}

@Article{Bullmore2009,
  author    = {E. Bullmore and O. Sporns},
  title     = {Complex brain networks: Graph theoretical analysis of structural and functional systems},
  journal   = {Nature Reviews Neuroscience},
  year      = {2009},
  volume    = {10},
  pages     = {186--198}
}

@Article{Luo2021, 
  author = {Luo, Y. and Cheng, Y. and Yang, H. and Liu, Z.},
  title = {Graph-Based Deep Learning for Medical Diagnosis and Analysis: Past, Present and Future}, 
  journal = {IEEE Access}, 
  year = {2021}, 
  volume = {9}, 
  pages = {145--165}
}

@misc{Algonauts2023Prep,
  author       = {Fales, G. and others},
  title        = {Algonauts 2023 Challenge: Data Preparation Pipeline},
  howpublished = {\url{https://github.com/gifale95/algonauts_2023}},
  year         = {2023},
  note         = {Accessed: 2025-01-15}
}

@article{Lin2015COCO,
  author       = {T.-Y. Lin and M. Maire and S. Belongie and L. Bourdev 
                  and R. Girshick and J. Hays and P. Perona 
                  and D. Ramanan and C. L. Zitnick and P. Doll{\'a}r},
  title        = {Microsoft COCO: Common Objects in Context},
  journal      = {arXiv preprint arXiv:1405.0312},
  year         = {2015}
}

@InProceedings{You2020GraphCL,
  author    = {Y. You and T. Chen and Y. Sui and Y. Shen and Z. Wang and Z. Shen},
  title     = {Graph Contrastive Learning with Augmentations},
  booktitle = {Proc. Adv. Neural Inf. Process. Syst. (NeurIPS)},
  year      = {2020},
  pages     = {5812--5823}
}

@article{Hutchison2013Dynamic,
  author    = {Hutchison, R. Matthew and Womelsdorf, Thomas and Allen, Elena A. and Bandettini, Peter A. and Calhoun, Vince D. and Corbetta, Maurizio and Della Penna, Stefania and Duyn, Jeff H. and Glover, Gary H. and Gonzalez-Castillo, Javier and Handwerker, Daniel A. and Keilholz, Shella and Kiviniemi, Vesa and Leopold, David A. and de Pasquale, Francesca and Sporns, Olaf and Walter, Martin and Chang, Catherine},
  title     = {Dynamic functional connectivity: Promise, issues, and interpretations},
  journal   = {NeuroImage},
  volume    = {80},
  pages     = {360-378},
  year      = {2013},
  doi       = {10.1016/j.neuroimage.2013.05.079}
}

@article{Baldassano2017Story,
  author    = {Baldassano, Christopher and Chen, Janice and Zadbood, Asieh and Pillow, Jonathan W. and Hasson, Uri and Norman, Kenneth A.},
  title     = {Discovering event structure in continuous narrative perception and memory},
  journal   = {Neuron},
  volume    = {95},
  number    = {3},
  pages     = {709-721.e5},
  year      = {2017},
  doi       = {10.1016/j.neuron.2017.06.041}
}

@article{Haxby2020Hyper,
  author    = {Haxby, James V. and Guntupalli, Jyothi S. and Nastase, Samuel A. and Feilong, Ma Feilong and Halchenko, Yaroslav O. and Connolly, Andrew C. and Ramadge, Peter J. and Gobbini, M. I. and Hanke, Michael},
  title     = {Hyperalignment: Modeling shared information encoded in idiosyncratic cortical topographies},
  journal   = {eLife},
  volume    = {9},
  pages     = {e56601},
  year      = {2020},
  doi       = {10.7554/eLife.56601}
}

@Article{Murphy2009GSR,
  author    = {K. Murphy and R. Birn and P. Bandettini},
  title     = {The impact of global signal regression on resting state correlations: Are anti-correlated networks introduced?},
  journal   = {NeuroImage},
  year      = {2009},
  volume    = {44},
  number    = {3},
  pages     = {893--905},
  doi       = {10.1016/j.neuroimage.2008.09.036}
}

@inproceedings{simonyan2014deep,
  title        = {Deep Inside Convolutional Networks: Visualising Image Classification Models and Saliency Maps},
  author       = {Simonyan, Karen and Vedaldi, Andrea and Zisserman, Andrew},
  booktitle    = {International Conference on Learning Representations (ICLR) Workshop},
  year         = {2014}
}

@inproceedings{shrikumar2017learning,
  title        = {Learning Important Features Through Propagating Activation Differences},
  author       = {Shrikumar, Avanti and Greenside, Peyton and Kundaje, Anshul},
  booktitle    = {Proceedings of the 34th International Conference on Machine Learning (ICML)},
  pages        = {3145--3153},
  year         = {2017}
}
\end{document}